\documentclass[10pt,twocolumn,letterpaper]{article}
\usepackage{iccv}
\usepackage{times}
\usepackage{epsfig}
\usepackage{graphicx}
\usepackage{amsmath}
\usepackage{amssymb}
\usepackage{booktabs}
\usepackage[ruled,vlined]{algorithm2e}
\usepackage{algorithmic}
\usepackage{multirow}
\usepackage{diagbox}
\usepackage{float}
\usepackage[english]{babel}
\usepackage{authblk}
\makeatletter
\renewcommand\AB@affilsepx{\quad\protect\Affilfont}
\makeatother

\usepackage[breaklinks=true,bookmarks=false]{hyperref}

\iccvfinalcopy 

\def\iccvPaperID{0056} 

\ificcvfinal\fi

\begin{document}

\title{Towards Class-wise Robustness Analysis}

\author{Tejaswini Medi$^{1}$, Julia Grabinski$^{1,2}$, and Margret Keuper$^{1,3}$\vspace{0.05cm}\\
$^1$ Visual Computing, University of Siegen\\
$^2$ Fraunhofer ITWM, Kaiserslautern and IMLA, University of Offenburg\\
$^3$ Max Planck Institute for Informatics, Saarland Informatics Campus}

\maketitle

\ificcvfinal\thispagestyle{empty}\fi
\begin{abstract}
While being very successful in solving many downstream tasks, the application of deep neural networks is limited in real-life scenarios because of their susceptibility to domain shifts such as common corruptions, and adversarial attacks. The existence of adversarial examples and data corruption significantly reduces the performance of deep classification models. Researchers have made strides in developing robust neural architectures to bolster decisions of deep classifiers. However, most of these works rely on effective adversarial training methods, and predominantly focus on overall model robustness, disregarding class-wise differences in robustness, which are critical. Exploiting weakly robust classes is a potential avenue for attackers to fool the image recognition models. Therefore, this study investigates class-to-class biases across adversarially trained robust classification models to understand their latent space structures and analyze their strong and weak class-wise properties. We further assess the robustness of classes against common corruptions and adversarial attacks, recognizing that class vulnerability extends beyond the number of correct classifications for a specific class. We find that the number of false positives of classes as specific target classes significantly impacts their vulnerability to attacks. Through our analysis on the Class False Positive Score, we assess a fair evaluation of how susceptible each class is to misclassification.


\end{abstract}

\section{Introduction}
Convolutional neural networks (CNNs) have achieved widespread success in various vision applications including image classification~\cite{he2016deep,huang2017densely,zagoruyko2016wide}, image segmentation\cite{Lin2020InteractiveIS}, and object detection~\cite{szegedy2013deep}. Nonetheless, the existence of adversarial examples~\cite{goodfellow2015explaining,pgd,croce2021mind, agnihotri2023cospgd} and common corruptions~\cite{hendrycks2019robustness} like blurring, zooming, or Gaussian noise, poses challenges in their real-world deployment. Extensive efforts have been devoted to defending against adversarial attacks and enhancing model generalization~\cite{Yuan2017AdversarialEA}. Adversarial training has emerged as a prominent defense technique to improve the robustness of classification models ~\cite{pmlr-v80-athalye18a,goodfellow2015explaining}. Prior works have analyzed adversarial training from different perspectives including robust optimization~\cite{pmlr-v97-wang19i}, robust generalization~\cite{raghunathan2019adversarial,grabinski2022aliasing,grabinski2022aliasing_aaai}, training strategy~\cite{pmlr-v97-zhang19p,pang2020boosting,Wang2020ImprovingAR} and neural architecture~\cite{grabinski2022frequencylowcut, grabinski2022robust, grabinski2023fix, lukasik2023improving, grabinski2024large, jung2023neural}. Howbeit, all these previous works have concentrated on improving the overall model robustness, neglecting the discrepancies in the robustness of individual classes. This imbalance in class-wise robustness can be exploited by attackers, who may target less robust classes. Therefore, a comprehensive understanding of adversarial training on class-wise robustness is crucial for improving the robustness of classification models in a meaningful way.
\par
Recently a few studies have emphasized class-wise robustness disparity in adversarial training ~\cite{DBLP:conf/kdd/0001KJW021,Benz2020RobustnessMB}. However, their focus has been limited to comparing class-wise robust accuracy deviations to identify the vulnerable classes. While this is important, analyzing class-to-class biases is equally crucial for gaining insights into the latent space of robust models. Specifically, understanding which class labels are assigned erroneously or which classes are predominantly confused is essential. Therefore, we conduct a comprehensive study of class-wise robust accuracies with particular emphasis on false positives in class-wise misclassifications, to improve the understanding of class-wise biases.
\par

\section{Background: Network Evaluation Methods}
When introducing new models for image classification tasks, network performance is typically evaluated in terms of accuracy \cite{he2016deep,zagoruyko2016wide,huang2017densely}. Such evaluation is most important, as we do not need a network that classifies randomly and wrongly. Additionally, the evaluation of the network's robustness received more popularity in the last few years. Hence, a variety of robustness measures has been proposed \cite{ho2021estimating,goodfellow2015explaining}. On the one hand,  common corruptions \cite{hendrycks2019robustness} have been introduced, which incorporate natural and system noise that can lead to misclassifications in the classification systems. On the other hand, adversarial attacks gained a lot of popularity to evaluate the network's vulnerabilities. In consequence, a variety of attacks, \eg \cite{goodfellow2015explaining,pgd,croce2021mind} along with their defenses have been proposed \cite{goodfellow2015explaining,pmlr-v97-zhang19p}. The \emph{robust accuracy} of a model is thereby usually defined as the model's accuracy under a specific adversarial attack or corruption. 
Thus, most of these studies 
focus on improving the overall robust accuracy of models under attacks or when facing corruptions. A few recent works further investigated the class biases in model accuracy and robustness, arguing for a fair training process that allows classifying all classes about equally well ~\cite{wei2023cfa,DBLP:conf/kdd/0001KJW021,Medi_2025_WACV}. These works also showed that adversarial training seems to amplify class-wise biases in model accuracy. Yet, only little effort has been devoted to studying which classes pre-dominantly attract incorrectly classified samples. To identify such classes, we study in this work the Class False Positive Score. We further argue that this perspective provides interesting insights into the model behavior and potentially allows to improve our understanding of the model's latent space and vulnerability.

\subsection{Evaluatiuon Metrics}
In this study, we calculate the \textbf{Class False Positive Score (CFPS)} to assess the vulnerability of each class $c_j$ towards misclassifications with $j\in \{1, \dots, C\}$ in the classification model. To calculate the CFPS for a specific class, we calculate the number of misclassifications where samples from other classes, \ie samples $x_i$ from the test set $\{ x_i\}_{i=1}^N$ of size $N$ with labels $y_i \in \{ c_j\}_{j=1}^C$  are incorrectly classified by model $f_{\theta}$ as this particular class, \ie the cardinality of $\{x_i|f_\theta (x_i)=c_j, y_i \neq c_j \}$. We then divide this count by the total number of misclassifications across all classes,
\begin{equation}
\mathrm{CFPS}(c_j) = \frac{|\{x_i|f_\theta (x_i)=c_j, y_i \neq c_j \}|}{|\{x_i|f_\theta (x_i) \neq y_i \}|}.
\end{equation}
A higher CFPS for a class indicates that it is more susceptible to being mistakenly assigned to samples from other classes by the model. The classes that are most likely mistaken as other classes have a high chance of manipulation by attackers, which impacts the overall reliability and security of a classification model. This enables us to focus on improving the robustness of these vulnerable classes.

The CFPS is complementary to the class-wise accuracy (CWA), which has been predominantly used in previous works such as \cite{DBLP:conf/kdd/0001KJW021,Benz2020RobustnessMB}, which is defined as
\begin{align}
&\mathrm{CWA}(c_j) = &&
{|\{x_i|f_\theta (x_i)=c_j, y_i = c_j \}|} /N \nonumber\\
&&+ &{|\{x_i|f_\theta (x_i)\neq c_j, y_i \neq c_j \}|}/N.
\end{align}
When evaluated under attack or corruption, we refer to these metrics as \emph{robust accuracy} and \emph{robust CFPS}, respectively.

\section{Experiments}
\paragraph{Class-Wise Accuracy Analysis.}
To carry out our experiments, we utilize the CIFAR-10 dataset, a simple and widely used benchmark for image classification tasks~\cite{krizhevsky2009learning}. For robustness evaluations, we employ adversarially trained robust models from a standardized adversarial robustness benchmark~\cite{grabinski2022robust,croce2020robustbench}. For our analysis, we select standard classification models like ResNet-18 and ResNet-50 from the ResNet family~\cite{he2016deep}, DenseNet-169~\cite{huang2017densely}, PreActResNet-18\cite{he2016identity}, WideResNet-70-16~\cite{zagoruyko2016wide} and a recent foundation model, DINOv2\cite{oquab2023dinov2}. The ten classes of CIFAR-10, namely 'airplane', 'automobile', 'bird', 'cat', 'deer', 'dog', 'frog', 'horse', 'ship', and 'truck', are denoted as C1 to C10 respectively in the following sections of the work.
 \par
 \begin{figure}[ht!]
  \centering
  \includegraphics[width=0.4\textwidth]{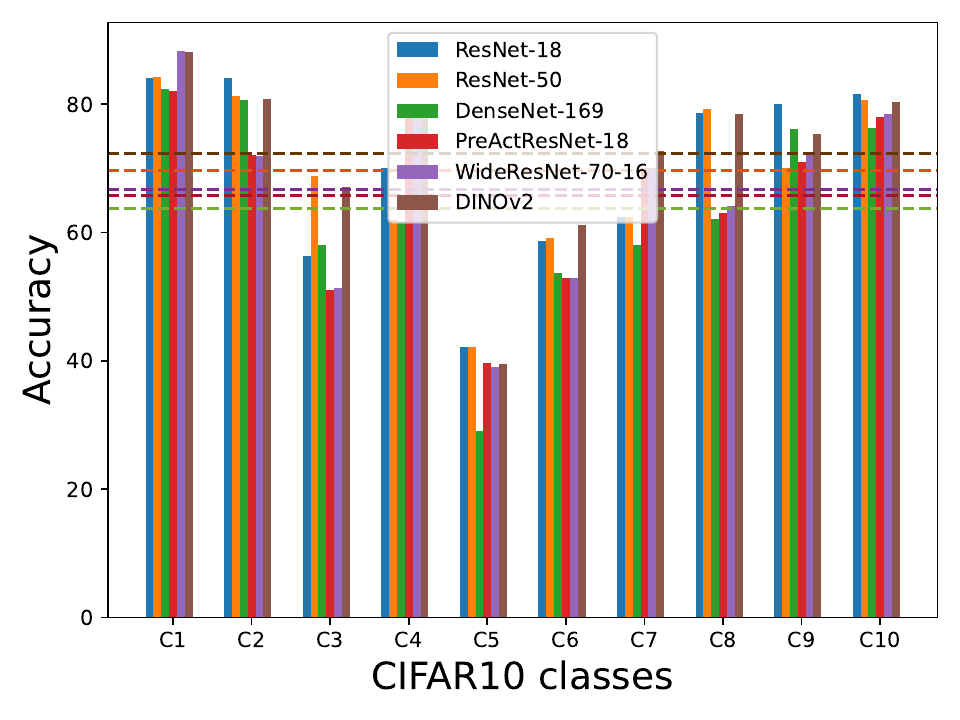}
\caption{Class-wise accuracies of CIFAR10 across different robust model architectures. The horizontal lines in the figure depict the average overall accuracy of respective adversarially trained robust models.}
\label{fig:robust}
\end{figure} 
Figure \ref{fig:robust} illustrates the class-wise accuracy of the aforementioned architectures when evaluated on clean validation samples from the CIFAR10 dataset. The overall accuracy of each model allows us to categorize classes into two groups: strong classes and weak classes. This categorization is based on whether a class exhibits a class-wise accuracy above or below the average overall accuracy of the model. Notably, we observe that classes C3, C5, and C6, corresponding to 'bird', 'deer', and 'dog' respectively, fall into the category of weak classes. This determination is made due to their relatively lower accuracy compared to the other classes across different robust models. It is important to emphasize that this pattern remains consistent regardless of the specific architecture employed for training the robust models. However, class-wise accuracy evaluation alone does not allow us to conclude that weaker classes are the primary source of confusion in model classification. Some classes may exhibit high accuracy yet still be frequently misclassified as other classes. Attackers can leverage this information to enhance the confusion in classifiers.


\par
In the context of class-wise robust analysis, previous research has commonly identified the weak robust classes
~\cite{DBLP:conf/kdd/0001KJW021,Benz2020RobustnessMB}. These determinations were often made by calculating the class-wise accuracy deviations with respect to the overall model accuracy or strong class accuracy but they failed to see a common pattern of weak classes under the influence of common corruptions and attacks. Moreover, such approaches may introduce biases, as the perceived weakness of a class could be affected by overall model accuracy or the performance of the strongest class. To ensure fairness and impartiality in our evaluation, we evaluate a metric called the Class False Positive Score, shortly CFPS. This metric focuses on model misclassifications among the classes independently of overall accuracy, enabling a comprehensive analysis of class-to-class biases exhibited by the models. 
\par
\begin{figure}[ht!]
  \centering
  \includegraphics[width=0.4\textwidth]{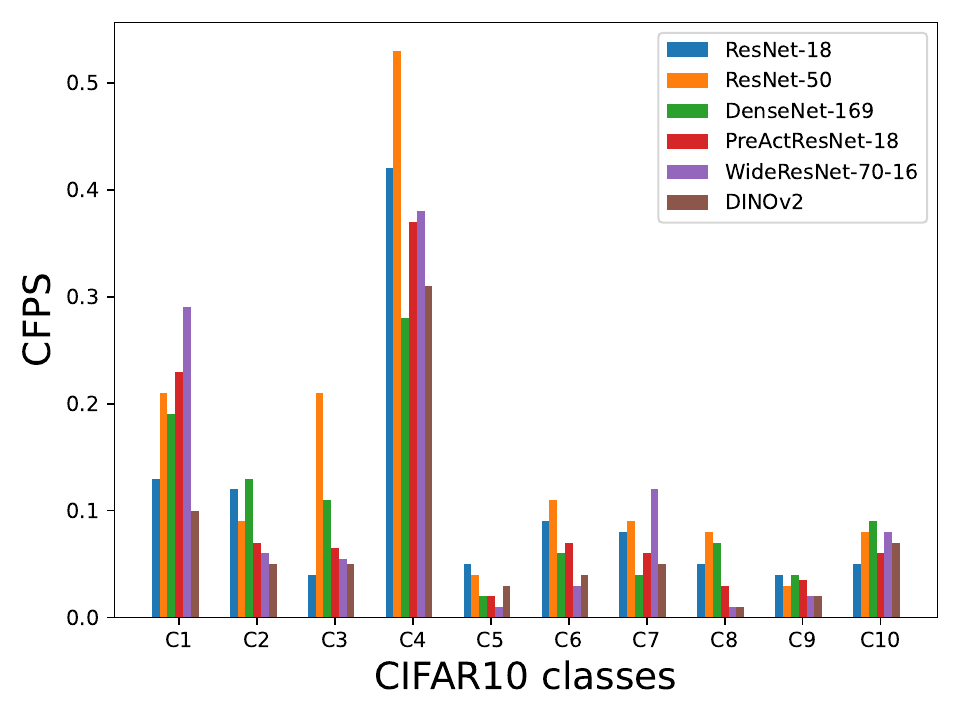}
\caption{Class-wise CFPS of CIFAR10 across different robust model architectures.}
\label{fig:MCS}
\end{figure}
\begin{figure*}[ht]
  \centering
  \includegraphics[width=\textwidth]{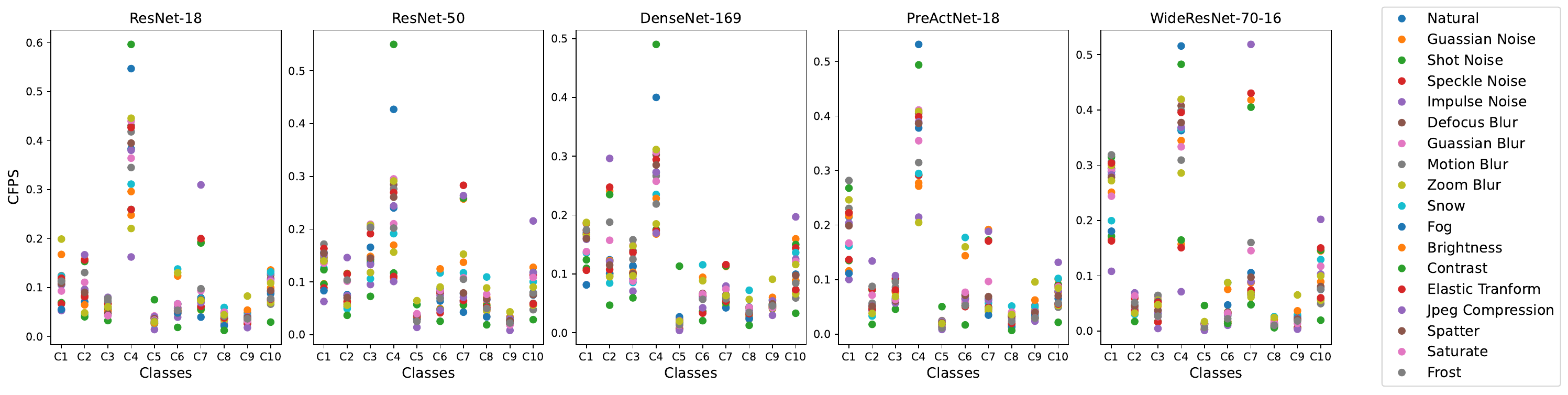}
\caption{Class-wise robust accuracies(top) and robust CFPSs (bottom) across different model architectures under corruptions. Robust accuracies are presented in fractions. Some classes with reasonably high robust accuracies tend to easily attract false positives and are thus overall more vulnerable than expected.}
\label{fig:MCS_C}
\end{figure*}

\noindent\textbf{Class False Positive Score. } 
Our evaluation of the CFPS for all classes of CIFAR-10 across different neural architectures is presented in Figure \ref{fig:MCS}. The results clearly demonstrate that the CFPS for classes C1, and C4 are comparatively higher, indicating these classes are highly susceptible to misclassifications. Conversely, the CFPS for the previously discussed weak classes C3, C5, and C6 is lower even though their class-wise accuracies are the least, suggesting relatively few samples are misclassified into these classes than into other classes. This finding underscores the importance of utilizing the CFPS metric, as it provides a more comprehensive and informative assessment of class-wise vulnerability to misclassifications and helps advance our understanding of the class-wise behavior of models in image classification tasks.
\vspace{-1em}
\paragraph{Class-Wise Robustness Analysis Under Common Corruptions.}
We assess the consistency of class-wise robust classification accuracies and CFPSs with commonly corrupted sample types on the CIFAR10-C dataset~\cite{hendrycks2019benchmarking}. Figure \ref{fig:MCS_C} presents class-wise robust accuracies and CFPSs of the adversarially trained aforementioned models across various corruption types. Interestingly, weak classes still consistently exhibit the lowest robust classification accuracy even with the inclusion of common corruptions, while C4 ('cat') still maintains the highest CFPS. i.e., the class vulnerabilities to misclassifications remained constant after the addition of common corruptions but the magnitude of vulnerability varies.

\noindent\textbf{ Which classes are more vulnerable to adversarial attacks, weak or highly misclassified?}
Experiments have revealed that the two indicators of the class-wise performance of the model (accuracy and CFPS) point to the distinct properties of the classes. A crucial question here is whether weak classes based on the least class-wise accuracy or those that are mostly misclassified as others are more susceptible to adversarial attacks. Therefore, we further investigate the influence of adversarial attacks~\cite{madry2019deep,goodfellow2015explaining} on class-wise robustness and also evaluate the most likely targetable class under the influence of attacks. We consider PGD attack~\cite{madry2019deep} using ResNet-50 \cite{chen2020adversarial} for this experiment.

\begin{figure*}[ht]
  \centering
  \includegraphics[width=\textwidth]{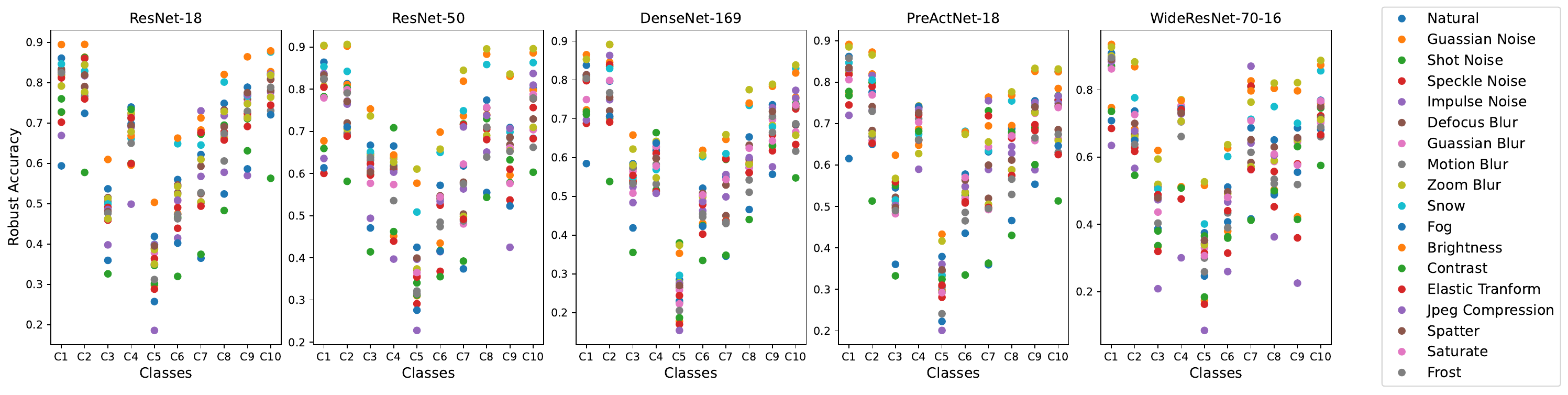}
\end{figure*} 

\begin{figure}[ht!]
  \centering
  \includegraphics[width =0.28\textwidth, height=0.26\textwidth]{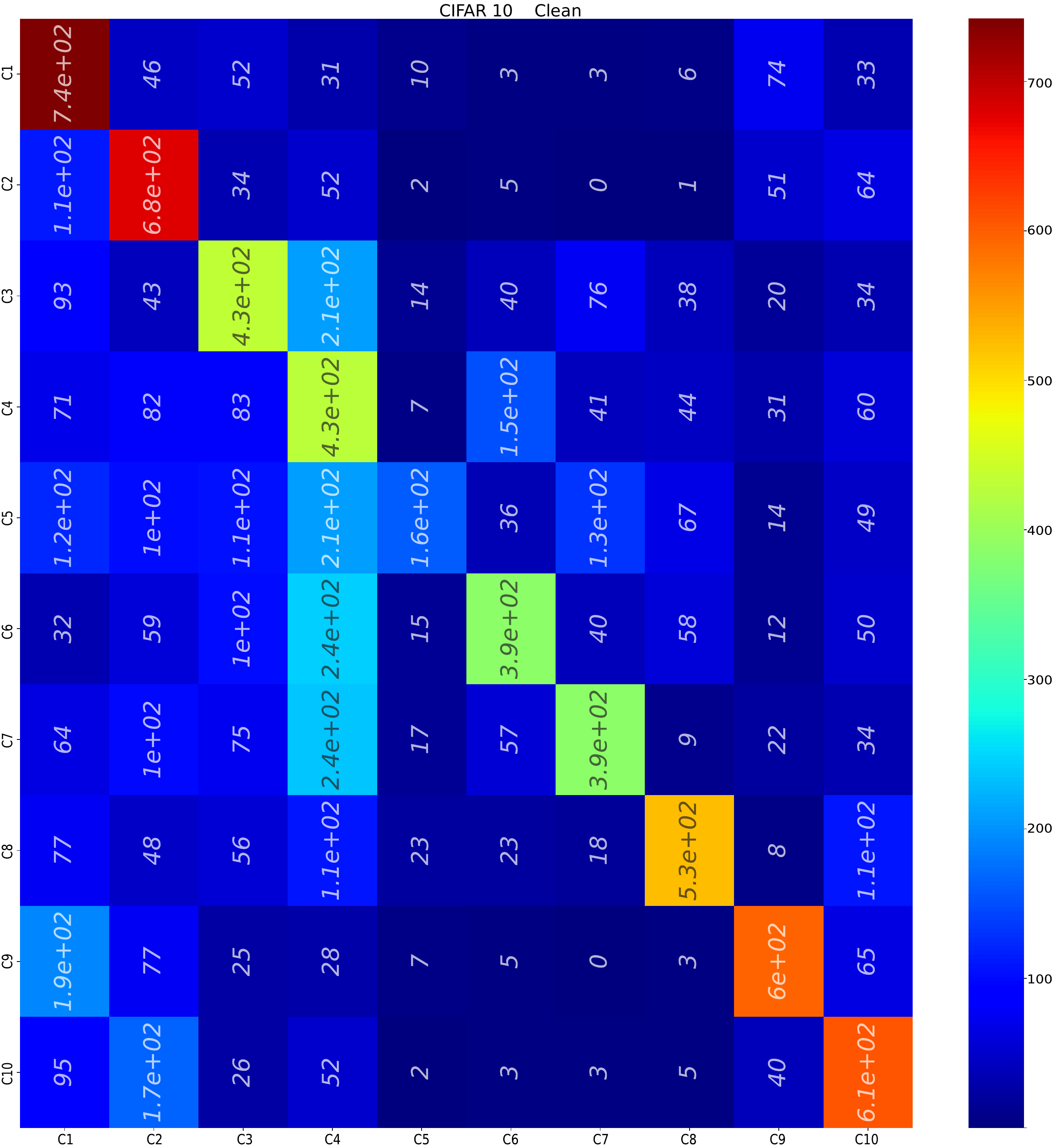}
\caption{Confusion Matrix defining ground truth (vertical axis) versus predictions(horizontal axis) under PGD attack.}
\label{FIG.ATTACK}
\end{figure}

Figure \ref{FIG.ATTACK} displays the confusion matrix depicting ground truth classes (vertical axis) versus the average of predicted classes over aforementioned models (horizontal axis) after subjecting to PGD attack with $\epsilon = 8/255$ and 20 attack steps. Following the heatmap color, the diagonal elements with the brightest blue shade indicate the lowest number of correct classifications per class and red color indicates the highest. Notably, the class C5 (deer) exhibits the lowest number of correct classifications, implying its vulnerability after subjecting the models to an adversarial attack. Furthermore, an observable pattern is the brightest vertical line aligned with the class C4 (cat) indicating that a significant portion of other classes is being misclassified as this class.
\par
\begin{figure}[ht!]
  \centering
  \includegraphics[width=0.32\textwidth]{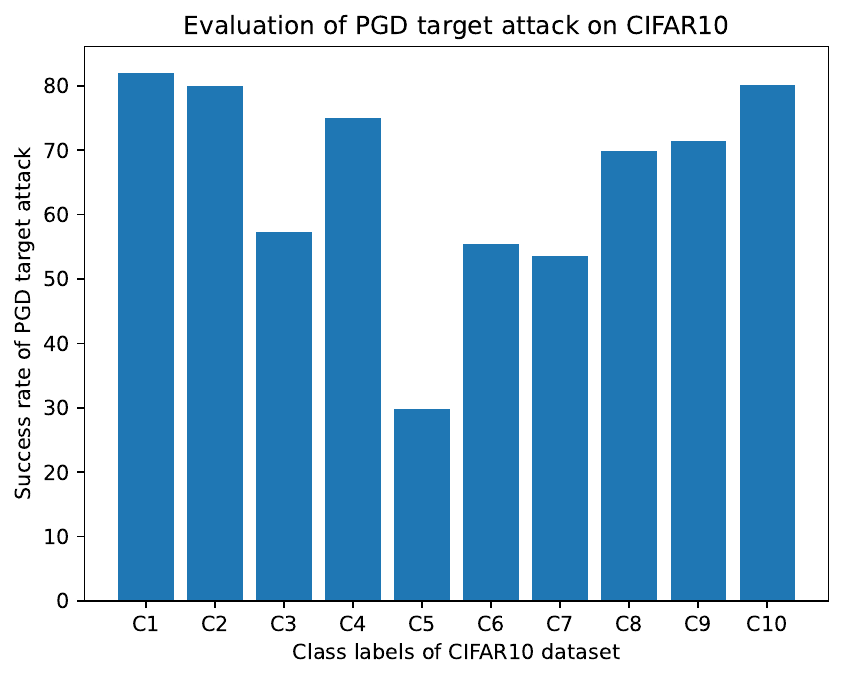}
   \caption{Evaluation of PGD target attack for all classes of CIFAR10 dataset using success rate.}
  \label{FIG.SR}
\end{figure}
\par

We further evaluate the success rate of PGD-targeted attacks using ResNet-50 \cite{chen2020adversarial} with $\epsilon = 2/255$ and 20 attack steps on the CIFAR10 dataset. Figure \ref{FIG.SR} shows the success rate evaluations of all target classes. The success rate is generally defined as the percentage of misclassifications tricked by the classification model under a targeted attack in the desired target class. We achieve a higher success rate of attack for the target "cat" than that of "deer". This illustrates that the "cat" class is more vulnerable to targeted attacks than "deer".

\section{Discussion}
The evaluations demonstrate that assessing the class-wise properties of a classification model requires considering both class-wise robust accuracy and CFPS. While the robust accuracy provides insights into a class vulnerability during 
adversarial attacks and corruptions, it does not necessarily reflect its class-wise susceptibility to misclassifications. 
It is crucial to identify the class C5 (deer) with the lowest robust accuracy, as it signifies the most vulnerable target for attacks. However, this vulnerability may not translate similarly when facing targeted adversarial scenarios. By examining the CFPS, we gain valuable information on this, for example the class "cat" as the most likely to be misclassified into (for CIFAR10). 
A potential reason is that class C4 (cat) is usually considered a rather difficult class because of the large intra-class variance in cat images. As a result, the label "cat" might tend to form a rather complex decision space, such that decision boundaries to this label can be easily reached from almost anywhere in the latent space. While this is a specific example on a specific dataset, we assume that similar biases exist across different datasets and models. Models are trained to reach high classification accuracies on potentially difficult classes while it is particularly easy for attackers to fool these models to misclassify other (potentially easy) samples into these classes.

\section{Conclusion}
In summary, this work studies both class-wise accuracy and class-wise false positives of classes to gain a comprehensive understanding of class-wise vulnerabilities and class-biases present in robust models, empowering us to develop more resilient defenses against potential attacks and corruptions or, at least, to better understand the behavior of our models under domain shifts. 


\newpage
{\small
\bibliographystyle{ieee_fullname}
\bibliography{egbib}
}
\newpage
\clearpage
\maketitle
\ificcvfinal\thispagestyle{empty}\fi

\raggedbottom  

This Supplementary Material provides additional details of our work.

\appendix
\addcontentsline{toc}{section}{Appendices}

\section{PGD Targeted attack on CIFAR10}
\begin{figure}[ht!]
  \caption{Evaluation of PGD target attack for all classes of CIFAR10 dataset using success rate.}
  \centering
  \includegraphics[width=0.4\textwidth]{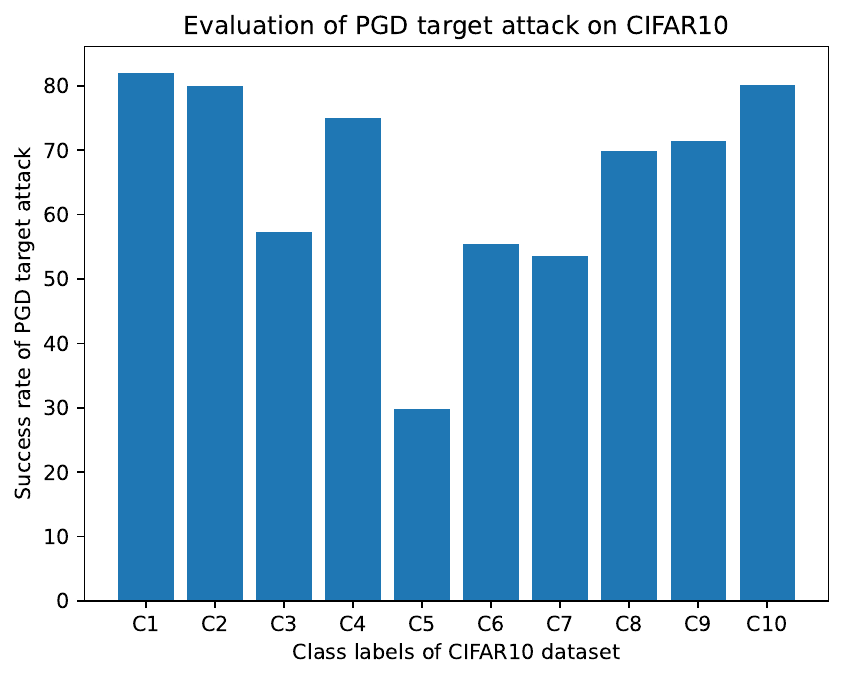}
  \label{FIG_bar}
\end{figure}

We evaluate the success rate of PGD-targeted attacks using PreActResNet-18 \cite{he2016identity} with $\epsilon = 2/255$ and 20 attack steps on the CIFAR10 dataset. Figure \ref{FIG_bar} shows the success rate evaluations of all target classes. The success rate is generally defined as the percentage of misclassifications tricked by the classification model under a targeted attack in the desired target class. We achieved a higher success rate of attack for the target C4 (cat) than for C5 (deer). This illustrates that the C4 class is more vulnerable to targeted attacks than C5.

\vspace{0.3cm}  
\vspace{0.3cm}
\section{Similarities Between Different Model Predictions} 
We evaluate class-wise robustness measures across different model architectures. For our analysis, we choose ResNet-18 and ResNet-50 from the ResNet family \cite{he2016deep}, DenseNet-169 \cite{huang2017densely}, PreActResNet-18 \cite{he2016identity}, and WideResNet-70-16 \cite{zagoruyko2016wide}. 

\begin{figure}[ht!]
  \caption{Cosine similarities between the class-wise predictions of different models}
  \centering
  \includegraphics[width=0.4\textwidth]{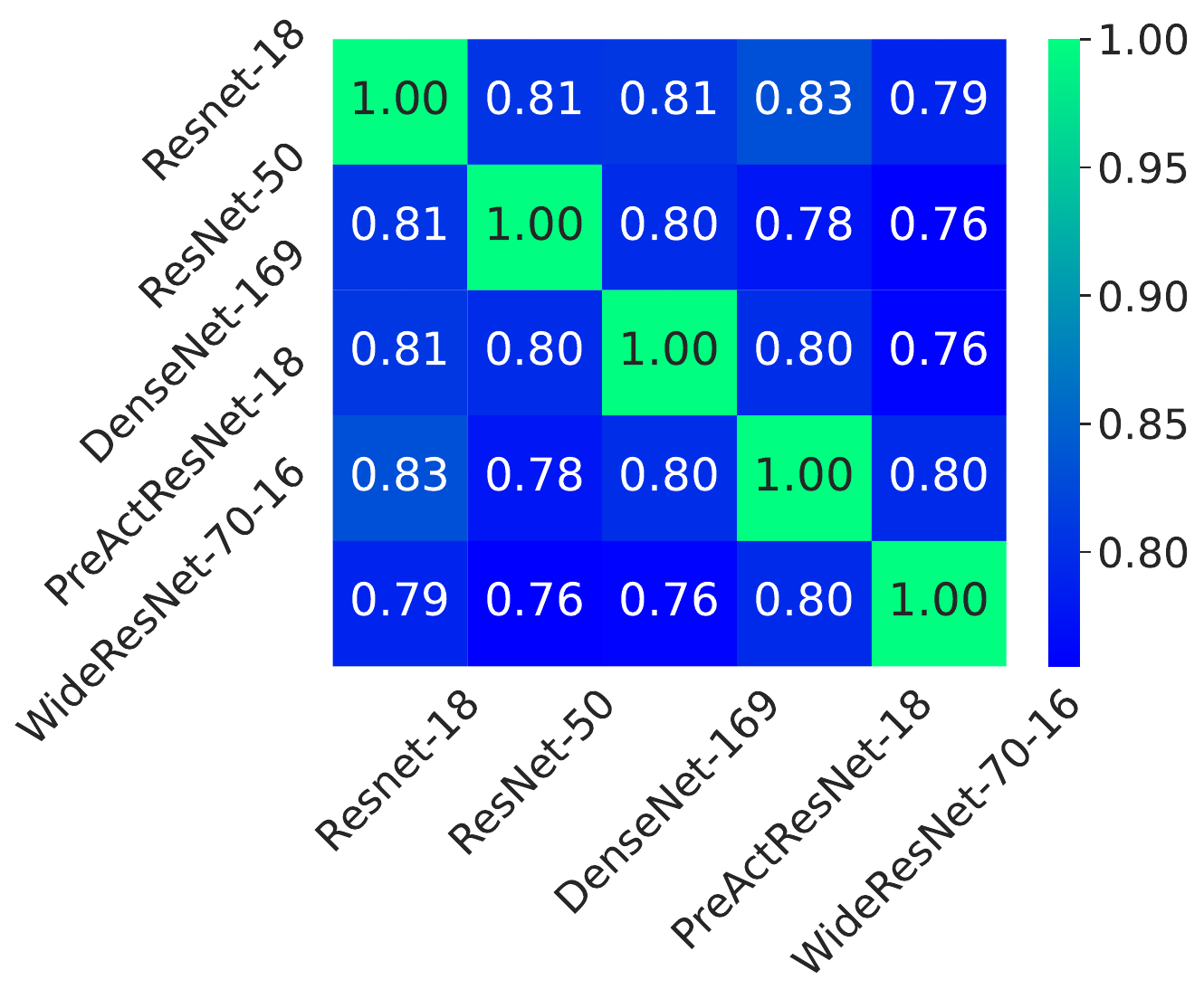}
  \label{FIG.Correlation}
\end{figure}

Understanding the correlation or similarities between the different model predictions is crucial to ensure the reliability and generalizability of these models. Figure \ref{FIG.Correlation} presents the cosine similarities between class-wise predictions across different models. 

The cosine similarity values are in the range of 0.76 to 0.83 across the aforementioned architectures, which indicates that the similarity between the predictions of the aforementioned models is high.

\vspace{0.3cm}  

\vspace{0.3cm} 

\section{Robust Model Details}
All the models for our work are selected from standardized adversarial robustness benchmark ~\cite{grabinski2022robust,croce2020robustbench}. For consistency and comparability, we adopt the same adversarial training approach as they proposed.

\raggedright 

\sloppy  

\end{document}